\documentclass[10pt,conference]{IEEEtran}
\IEEEoverridecommandlockouts
\usepackage{amsmath,amssymb,amsfonts}
\usepackage[ruled,linesnumbered]{algorithm2e}
\usepackage{graphicx}
\usepackage{textcomp}
\usepackage{float}
\usepackage{booktabs}
\usepackage{multirow}
\usepackage{array}
\usepackage{amssymb}
\usepackage{paralist}
\usepackage{xcolor}
\def\BibTeX{{\rm B\kern-.05em{\sc i\kern-.025em b}\kern-.08em
    T\kern-.1667em\lower.7ex\hbox{E}\kern-.125emX}}

\begin{document}

\title{LSAQ: Layer-Specific Adaptive Quantization for Large Language Model Deployment
\thanks{*Corresponding author.}
}
\author{\centering
\IEEEauthorblockN{Binrui Zeng, Bin Ji, Xiaodong Liu{*}, Jie Yu{*}, Shasha Li, Jun Ma, Xiaopeng Li, }
\IEEEauthorblockN{Shangwen Wang, Xinran Hong, Yongtao Tang}
\IEEEauthorblockA{
College of Computer Science and Technology,
National University of Defense Technology,
Changsha, China\\
\{zengbinrui, jibin, liuxiaodong, yj, shashali, majun, xiaopengli, wangshangwen13, hongxinran, tangyt\}@nudt.edu.cn}
}

\maketitle

\begin{abstract}
% As large language models (LLMs) excel in tasks like creative writing and code generation, their deployment in data centers with high-end GPUs raises challenges related to privacy and connectivity. Deploying LLMs on edge devices, particularly those with consumer-grade GPUs, can address these issues by enabling faster responses and protecting user privacy. This paper proposes a layer-specific quantization and dynamic deployment method for LLMs, which ranks layer importance and adapts quantization strategies based on device resources. The method includes three modules: layer importance detection, device resource detection, and quantization strategy formulation. Experimental results show that our approach, LSAQ, significantly reduces model storage while maintaining performance, outperforming existing quantization methods on multiple LLMs.
As Large Language Models (LLMs) demonstrate exceptional performance across various domains, deploying LLMs on edge devices has emerged as a new trend. Quantization techniques, which reduce the size and memory requirements of LLMs, are effective for deploying LLMs on resource-limited edge devices. However, existing one-size-fits-all quantization methods often fail to dynamically adjust the memory requirements of LLMs, limiting their applications to practical edge devices with various computation resources. To tackle this issue, we propose \textbf{L}ayer-\textbf{S}pecific \textbf{A}daptive \textbf{Q}uantization (\textbf{LSAQ}), a system for adaptive quantization and dynamic deployment of LLMs based on layer importance. Specifically, LSAQ evaluates the importance of LLMs' neural layers by constructing top-\(k\) token sets from the inputs and outputs of each layer and calculating their Jaccard similarity. Based on layer importance, our system adaptively adjusts quantization strategies in real time according to the computation resource of edge devices, which applies higher quantization precision to layers with higher importance, and vice versa. {Experimental results show that LSAQ consistently outperforms the selected quantization baselines in terms of perplexity and zero-shot tasks. Additionally, it can devise appropriate quantization schemes for different usage scenarios to facilitate the deployment of LLMs.}
% {Experimental results on This approach significantly reduces the storage requirements of LLMs while maintaining model performance, enabling efficient deployment across diverse hardware platforms and usage scenarios.} 

\end{abstract}


\section{Introduction}
With the rapid advancement of Large Language Models (LLMs), they have demonstrated exceptional capabilities across a variety of tasks. Taking advantage of these capabilities, LLMs have been widely studied in multiple research domains like natural language processing \cite{qin2024largelanguagemodelsmeet}, code generation \cite{jiang2024surveylargelanguagemodels,li2024modeleditingllms4codefar}, finance \cite{zhao2024revolutionizing} and education \cite{wang2024large}. To harness LLMs for more intelligent and personalized services \cite{lyu2023llm} while ensuring data security and privacy, deploying LLMs on edge devices has emerged as a new trend. {However, LLMs' massive parameters pose significant challenges for efficiently deploying them on resource-limited edge devices. For example, Llama-2-13B \cite{touvron2023llama} requires about 25GB GPU memory when loading its weights. In contrast, a high-performance consumer-grade GPU such as the RTX 4090 has only 24 GB of memory, rendering it incapable of deploying Llama-2-13B directly.} To address this challenge, researchers explore LLM compression techniques such as quantization \cite{frantar2022gptq,dettmers2022llmint88bitmatrixmultiplication,lin2024duquantdistributingoutliersdual}, pruning \cite{ma2023llm,frantar2023sparsegpt,yang2024lacolargelanguagemodel}, distillation \cite{gu2024minillm,huang2022context}, and low-rank factorization \cite{xu2023tensorgpt} to reduce GPU memory requirements when deploying LLMs.

In the field of LLM compression, quantization plays a pivotal role. Its core mechanism involves converting high-precision floating-point weights or activation values included in an LLM into lower-precision numerical representations. This transformation significantly reduces the memory requirements of LLM deployment. Post-Training Quantization (PTQ), a widely studied quantization technique, is particularly valuable for deploying LLMs on resource-limited devices, such as personal computers. A number of PTQ methods have achieved great success, such as LLM.int8() \cite{dettmers2022llmint88bitmatrixmultiplication}, GPTQ \cite{frantar2022gptq}, and Omniquant \cite{shao2024omniquantomnidirectionallycalibratedquantization}.  

However, the majority of existing PTQ methods uniformly quantize across all neural layers of LLMs, neglecting the fact that these layers are of quite different importance \cite{men2024shortgpt}. For example, substantial redundancy exists in numerous LLMs' neural layers. This one-size-fits-all quantization approach also limits the ability to dynamically adjust the memory requirements when deploying LLMs to various edge devices with quite different computation resources. Layer-Wise Quantization (LWQ) \cite{dumitru2024layer} offers a potential solution to this problem by quantifying each layer based on layer importance, which is calculated by the cosine similarity between the input and output of the layer, or the distribution of weight outliers. Despite the success of LWQ, it exposes nonnegligible limitations, i.e., it cannot reflect the degree of semantic change directly and struggles to effectively distinguish layers with relatively uniform weight distributions. In light of this, it is crucial to explore more refined quantization strategies to enable differentiated treatment of LLM layers. \textcolor{black}{Additionally, techniques should be developed to dynamically adjust the quantization precision. This should be based on practical application scenarios to further optimize storage and inference efficiency, while maintaining LLM performance.}
%Additionally, techniques capable of dynamically adjusting the quantization precision based on practical application scenarios should be developed to further optimize the storage and inference efficiency while maintaining model performance.(split into two)

\begin{figure}[t]
    \centering
    \includegraphics[width=1\linewidth]{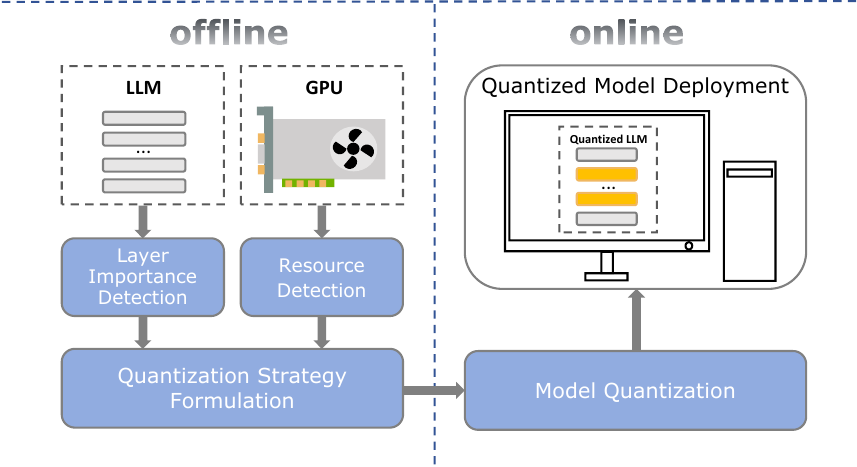}
    \caption{The framework of LSAQ. \textcolor{black}{It is composed of offline and online parts. In the offline part, the importance of each layer of the LLM is first obtained, and the available GPU resources at the current moment are detected simultaneously. Based on this, a quantization strategy is meticulously formulated. Subsequently, this quantization strategy is transmitted to the online part, where the model is quantized according to this strategy.}}
    \label{fig:framework}
    % \vspace{-0.5cm}
\end{figure}

To tackle the above issues, we propose LSAQ, a novel \textbf{L}ayer-\textbf{S}pecific \textbf{A}daptive \textbf{Q}uantization system tailored for the deployment of LLMs, as illustrated in Figure \ref{fig:framework}. LSAQ introduces a fine-grained layer importance evaluation mechanism by constructing top-\(k\) token sets corresponding to the inputs and outputs of each layer in LLMs. We propose to calculate the Jaccard similarity between these two sets as the indicator of layer importance. A higher Jaccard similarity implies greater similarity between the input and output token sets, indicating that the layer fails to significantly encode the semantics when processing the input. Motivated by the above fact, we regard such layers as less important. Next, our system adaptively adjusts the quantization strategy according to the available computation resource of an edge device, applying higher quantization precision to layers with greater importance and lower precision to less important layers. Additionally, our system is capable of maximizing the number of high-precision layers to preserve LLM performance, which enables the full usage of computation resources. Finally, the quantized LLMs are deployed to edge devices to handle various tasks. 
%Our LSAQ significantly reduces the requirements LLMs' computation resources while maintaining higher LLMs' performance. It also enables efficient deployment of LLMs across various edge devices and various usage scenarios on the same device. 

We conducted experiments on Llama models with parameter sizes ranging from 7B to 13B, and the results show that LSAQ outperformed existing same-granularity quantization method in accuracy across 75\% of the selected zero-shot tasks, and demonstrated superiority in average accuracy across all tasks. Additionally, in 90\% of the cases, LSAQ performed better than the same-granularity quantization method on the perplexity metric based on the WikiText2 \cite{merity2016pointer} dataset. Furthermore, we carried out deployment tests on various devices and found that LSAQ was able to select the optimal deployment scheme for the deployment of LLMs.

The contributions of this work can be summarized as follows:
(1) As far as we know, we are the first to use the Jaccard similarity to calculate the importance of LLMs' neural layers. Experimental results reveal its superiority to previous methods like cosine similarity \textcolor{black}{(See $\S$\ref{sec:main_results})};
(2) Based on layer importance, we propose a layer-specific quantization strategy, enabling us to dynamically deploy LLMs to various edge devices with different computation resources, meanwhile maintaining the LLMs' capabilities;
(3) \textcolor{black}{Compared to methods with the same quantization granularity, LLMs using LSAQ quantization to achieve better performance in terms of accuracy on zero-shot tasks and perplexity based on WikiText2.}

\section{Related Work}

\subsection{LLMs Quantization}
Quantization is a critical technique in the field of deep learning. It significantly reduces the model size by converting model parameters from high-precision formats (e.g., 32-bit or 16-bit floating-point numbers) to lower-precision formats (e.g., 8-bit or 4-bit integers), making it feasible to deploy LLMs on resource-limited edge devices.  

There are two typical approaches for LLM quantization: Quantization-Aware Training (QAT) and Post-Training Quantization (PTQ). QAT incorporates quantization operations during LLMs' training phase. It uses high-precision parameters for backpropagation and low-precision parameters for forward propagation, enabling LLMs to adapt to the changes introduced by quantization. In contrast, PTQ is applied after finishing model training. It reduces the representation precision of weights and activation values to reduce LLMs' size. Compared to QAT, PTQ is more suitable for scenarios requiring rapid LLM deployment on resource-limited devices.

Recently, PTQ has made significant advancements in reducing memory usage and improving computational efficiency for LLMs. The ZeroQuant \cite{yao2022zeroquant} technique supports INT8 symmetric quantization and allows different quantization levels to be set for the final model, adapting to diverse model and task requirements. The GPTQ \cite{frantar2022gptq} technique determines quantization parameters using calibration data, achieving the goal of reducing model size while maintaining accuracy. LLM.int8() \cite{dettmers2022llmint88bitmatrixmultiplication} improves quantization efficiency by handling activation outliers through mixed-precision decomposition. The SmoothQuant \cite{xiao2023smoothquant} technique introduces a mathematically equivalent per-channel scaling transformation to smooth activations and their corresponding weights across different channels, making the model more quantization-friendly. AWQ \cite{lin2024awq} optimizes the LLM quantization process by preserving significant weights and protecting prominent ones.

The development of these techniques has collectively facilitated the deployment and application of large language models in resource-limited environments, offering new insights and methods for optimizing and deploying LLMs.

\begin{figure*}[t]
    \centering
    \includegraphics[width=1\linewidth]{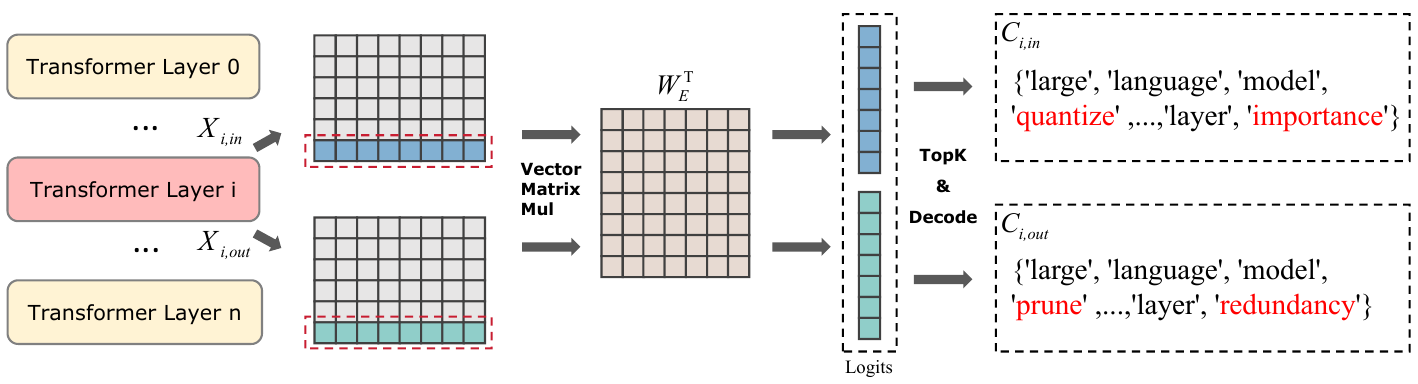}
    \caption{The process of constructing top-\(k\) token sets.}
    \label{fig:Jaccard}
\end{figure*}

\subsection{Layer Importance Assessment for LLMs}
In the field of research on LLMs, prior studies like shortGPT \cite{men2024shortgpt} have revealed that certain layers within these models may perform similar or redundant operations. These layers exhibit a high degree of functional overlap in the information processing pipeline, leading to redundancy in their contributions to the LLM’s final output. By thoroughly analyzing the contribution of each layer to the LLM, we can determine the relative importance of individual layers. This analysis not only helps identify layers with minimal impact on LLM performance but also provides a theoretical foundation for LLM compression.  

When exploring the importance of layers in LLMs, Men et al. \cite{men2024shortgpt} and Dumitru et al. \cite{dumitru2024change} have proposed metrics such as block influence and layer redundancy to define layer importance. Both studies share a core mechanism based on calculating the cosine similarity between the input and output of each layer. Cosine similarity measures the angular similarity between two non-zero vectors, where values closing to 1 indicate higher similarity, implying that the operations performed by the layer are more redundant or overlapping. In addition to using cosine similarity as a metric, Dumitru et al. \cite{dumitru2024layer} also estimate the layer importance by analyzing the number of weights significantly exceeding the average value within a layer. If a layer contains many weights with absolute values much larger than the average weight of that layer, it is likely to have a greater impact on the LLM's output.

By evaluating layer importance, the contribution of each layer to the overall functionality of the LLM can be quantitatively assessed. This approach provides a clear framework for analyzing and comparing the roles of different layers, facilitating the identification and optimization of key layers. As a result, it helps enhance the LLM’s efficiency and performance.

\section{Method}

\begin{algorithm*}[ht]
    \caption{Quantization Precision Allocation}
    \label{algo:qpa}
    \KwIn{$\mathcal{L}$: List of layer importance (in order of increasing importance), $\mathcal{M}_{ava.}$: Memory of currently available GPU}
    \KwOut{$\mathcal{L}_{fp16}$, $\mathcal{L}_{int8}$, $\mathcal{L}_{int4}$: Lists of layer's label corresponding to different precisions}

    $\mathcal{L}_{fp16}$, $\mathcal{L}_{int8}$, $\mathcal{L}_{int4}$ $\gets$ [\ ]\;

    \uIf{$\mathcal{M}_{ava.}$ $\geq$ Memory required for FP16 precision}{
        $\mathcal{L}_{fp16}$ = $\mathcal{L}$\;
    }\uElseIf{$\mathcal{M}_{ava.}$ $\geq$ Memory required for INT8 precision}{
        $\mathcal{L}_{int8}$ = $\mathcal{L}$\;
    }\uElseIf{$\mathcal{M}_{ava.}$ $\geq$ Memory required for INT4 precision}{
        set $len$ to the number of layers in LLM\;
        set $\mathcal{M}_{int8}$ to the difference between the $\mathcal{M}_{ava.}$ and memory required for INT4 precision\;
        % set $\mathcal{S}_{int8}$ to the quotient of $\mathcal{M}_{int8}$ and the savings by quantizing a single layer from INT8 to INT4\;
        set $\mathcal{S}_{8to4}$ to the memory saved by quantizing a single layer from INT8 to INT4 precision\;
        $\mathcal{N}_{int4}$ = $len$ - {$\mathcal{M}_{int8}$} / {$\mathcal{S}_{8to4}$}\; 
        $\mathcal{L}_{int4}$ = $\mathcal{L}[\ :\mathcal{N}_{int4}]$\;
        $\mathcal{L}_{int8}$ = $\mathcal{L}[\mathcal{N}_{int4}:\ $$len$$]$\;
    }\Else{
        pause quantization process until other processes release memory\;
    }
    
    \Return $\mathcal{L}_{fp16}$, $\mathcal{L}_{int8}$, $\mathcal{L}_{int4}$\;
    
\end{algorithm*}

This section introduces LSAQ, a system designed for \textbf{L}ayer-\textbf{S}pecific \textbf{A}daptive \textbf{Q}uantization and dynamic deployment of LLMs on edge devices equipped with consumer-level GPUs. As shown in Figure \ref{fig:framework}, our system is composed of both offline and online components. Due to variations in layer importance across different LLMs, as well as differences in GPU resources available on various edge devices or even on the same device under different conditions, the offline component must determine the most suitable quantization strategy based on these factors. During the online phase, the selected quantization strategy is applied to quantize the LLM to the corresponding bit widths. The quantized LLM is then deployed onto the target device, ensuring that the LLM can operate efficiently in resource-limited environments while maintaining a high level of accuracy.

\subsection{Layer Importance Detection Module}

The Layer Importance Detection (LID) module identifies the key layers that have the greatest impact on LLM performance by analyzing the structure and behavior of the LLM. This ensures that essential functions of the LLM are preserved during quantization. 
% To assess layer importance, we utilize intrinsic model features as metrics. 
% To characterize the influence of each layer on the final output of LLMs, ShortGPT introduces Block Influence (BI) as an evaluation metric for layer importance. Essentially, BI calculates the cosine similarity between each layer's input and output values. Layers with lower cosine similarity (indicating lower directional similarity) are considered to perform critical information processing within the model.

To assess the layer importance, we utilize intrinsic model features as metrics. We analyze each layer of the LLMs, extracting the hidden states of {the last token}, denoted as \(X_{i,in}\) and \(X_{i,out}\). These hidden states capture the encoded information for the last token in the input sequence at that particular layer. Next, we multiply these hidden states by the transpose of the matrix \(W_{E}\). {The matrix \(W_{E}\) serves as an input embedding matrix, which is capable of mapping high-dimensional input data into a lower-dimensional vector space, thereby extracting the key features of the hidden states.} We can directly obtain the matrix from the corresponding LLM. This multiplication operation produces a projection of each token across the vocabulary.

Following this, we sort the results of the projection in descending order. By selecting the top-\(k\) indices from the sorted results, we identified the top-\(k\) most probable tokens for the input and output of the current layer. Among them, the selection of the \(k\) value is based on the work of Li et al \cite{li2024pmet}. However, the specific value chosen for \(k\) has minimal impact on the overall trend. This is because, in reality, smaller \(k\) values are essentially subsets of larger \(k\) values. 

Finally, we convert these indices back into actual tokens, forming top-\(k\) token sets that represent the input and output of the layer. Figure \ref{fig:Jaccard} illustrates the detailed process of constructing the top-\(k\) token sets.
\begin{equation}
C_{i,in}=f_{topk}({X_{i,in}\cdot W_{E}^\top}).\label{eq1}
\end{equation}
\begin{equation}
C_{i,out}=f_{topk}({X_{i,out}\cdot W_{E}^\top}).\label{eq2}
\end{equation}

After obtaining the two sets \( C_{i,in} \) and \( C_{i,out} \) \textcolor{black}{through Equ.\eqref{eq1} and Equ.\eqref{eq2},} we calculate the Jaccard similarity \cite{niwattanakul2013using} between the two sets as a metric for assessing the importance of the layer. Jaccard similarity has been widely used in various fields (e.g., machine learning and text mining) to evaluate the similarity between sets of samples or features. This similarity is calculated based on the ratio of the size of the intersection of the above two sets to the size of their union. This metric is simple yet effective, providing a clear measure of the compositional similarity between the two top-\(k\) token sets.
\begin{equation}
I_{i}=1-J(C_{i,in},C_{i,out})=1-\frac{|C_{i,in}\cap C_{i,out}|}{|C_{i,in}\cup C_{i,out}|}.\label{eq3}
\end{equation}

According to Equ.\eqref{eq3}, we employ the Jaccard similarity to quantify the semantic transformation degree of each layer in LLMs. \textcolor{black}{Specifically, we calculate the Jaccard similarity between two top-\(k\) token sets derived from the input and output of each layer. Then, we transform this value by inversion and addition to obtain the importance metric \(I_{i}\) for each layer.} As the Jaccard similarity increases, the importance metric \(I_{i}\) decreases correspondingly. 
\begin{figure}[ht]
% \vspace{-0.4cm}
    \centering
    \includegraphics[width=1\linewidth]{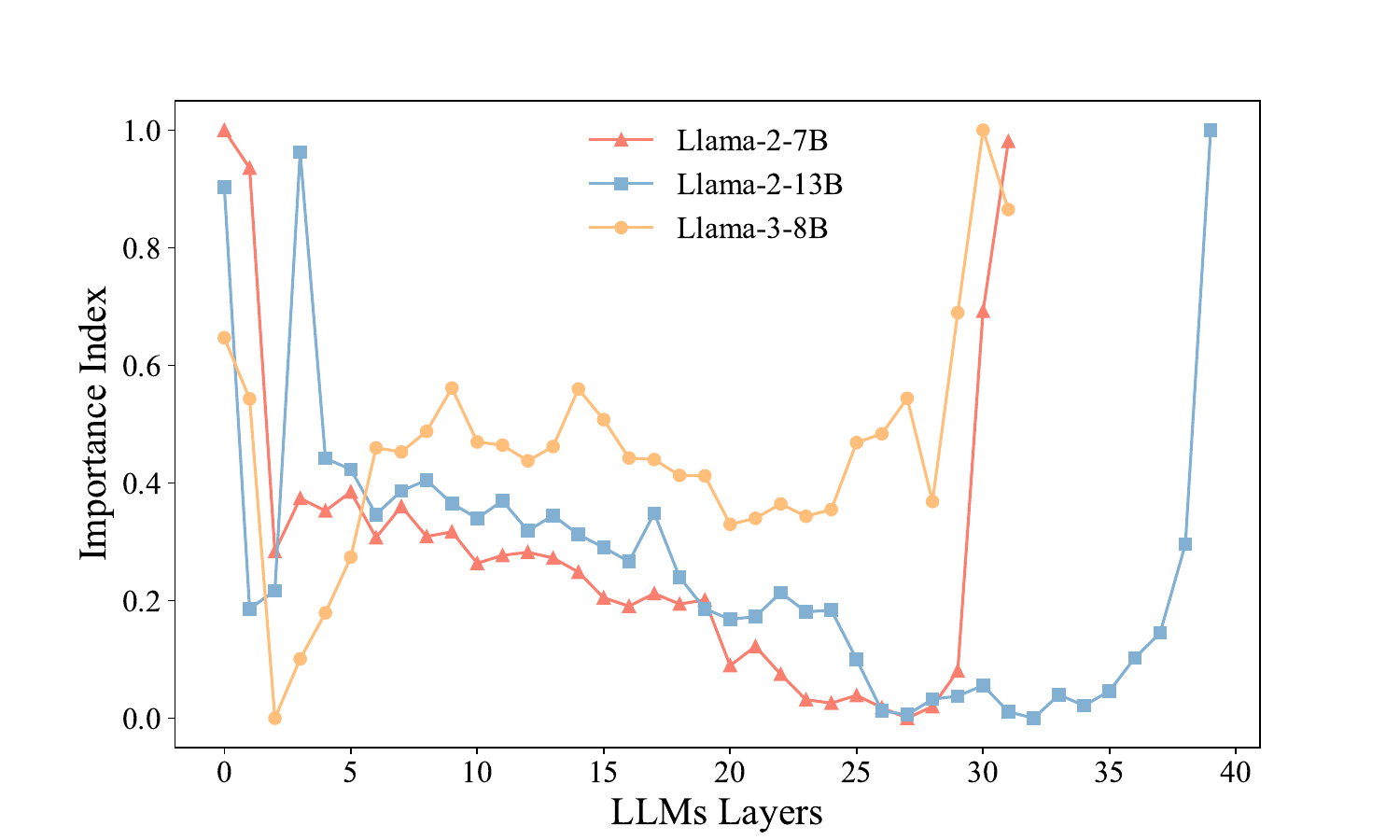}
    \caption{Importance of LLMs layer.}
    \label{fig:Importance Index}
\end{figure}

This observation indicates that a higher similarity between the two top-\(k\) token sets implies that the layer may have performed less significant semantic transformation on the input information. Consequently, it can be inferred that the layer's contribution to semantic transformation is relatively low, leading to a lower assessment of its importance. Through Figure \ref{fig:Importance Index}, we can visually observe the contribution of different layers to the overall performance of the Llama-2-7B/13B \cite{touvron2023llama} and Llama-3-8B \cite{dubey2024llama} within this evaluation framework.

Compared to directly using cosine similarity, our approach offers a more fine-grained analysis by directly mapping hidden states to the vocabulary. This method captures the relationship between each layer and semantic information more effectively, helping to reveal the features and aspects that each layer focuses on. As a result, it enables a more accurate assessment of the importance of the layer.

\subsection{Resource Detection Module}
Effective management of GPU resources is critical for the edge-side deployment of LLMs, as it directly impacts the feasibility of deployment as well as the subsequent performance and responsiveness of the LLMs. The Resource Detection (RD) module is designed to adapt to GPU devices of various scales and configurations, including scenarios of personal computers, servers, or cloud computing environments.  

Through flexible environment variable configurations and programmatic approaches, the RD module can automatically select the GPU device with the most available free memory. This capability ensures efficient loading and inference of quantized LLMs, optimizing their performance during deployment.

\subsection{Quantization Strategy Formulation Module}

The core of the Quantization Strategy Formulation (QSF) module lies in designing an optimal LLM quantization deployment plan based on the layer importance of the target LLM and the current GPU resource availability. This strategy aims to ensure that the quantized LLM minimizes accuracy loss while enabling seamless deployment and efficient inference execution. We present the complete pseudocode for QSF in Algorithm \ref{algo:qpa}.

Specifically, the module first assesses whether the LLM can be deployed directly without quantization or after quantization to INT8, based on the available GPU resources. In both deployment scenarios, the accuracy loss of the LLM is negligible. Furthermore, if the GPU resources are insufficient to support the two aforementioned types of precision models but can accommodate the INT4 quantization model, the module will apply INT4 quantization to the less important layers. The remaining layers will continue to use the INT8 quantization. This approach maximizes the proportion of INT8 quantized layers to maintain the accuracy of the LLM. Finally, if the GPU resources cannot meet any of the above conditions, the quantization task will be paused and wait for other processes to release the necessary resources.

% Specifically, this module first evaluates the relationship between GPU memory availability and the memory requirements of the LLM during loading and inference. If the GPU memory is sufficient, the LLM is deployed without quantization to maintain optimal performance. If the memory is insufficient for an unquantized LLM but sufficient for an INT8-quantized LLM, the entire LLM undergoes INT8 quantization, as the accuracy loss from INT8 quantization is negligible.

% Further, if the GPU memory is insufficient for an INT8-quantized LLM but can support an INT4-quantized LLM, the module applies INT4 quantization to layers with low importance while retaining INT8 quantization for the remaining layers. Since INT4 quantization has a greater impact on LLM accuracy, the strategy aims to maximize the proportion of INT8-quantized layers while ensuring that the quantized LLM's memory requirements remain within the GPU's capacity. Finally, if the GPU memory cannot accommodate even an INT4-quantized LLM, the quantization process will be paused until other processes release memory. 

This module aims to achieve an optimal balance between LLM accuracy and GPU resources, ensuring adaptability to various deployment environments.

\subsection{Model Quantization Module}
After determining the deployment strategy via QSF, this module performs per-channel quantization on the layers of LLMs that require quantization. Compared to the per-tensor quantization method, per-channel quantization assigns independent quantization parameters to each channel of the weight matrix. This approach allows the quantization process to more accurately capture the dynamic range of data in each channel, thereby reducing quantization errors and improving LLM accuracy. The quantization process can be mathematically represented as:
\begin{equation}
W_{i}^{INTn}=round(\frac{W_{i}^{FP16}}{s_{i}})\quad s.j.t. \quad s_{i} = \frac{\max\{|W_{i}|\}}{2^{n-1}-1},\label{eq4}
\end{equation}
where \( W \) denotes the weight matrix of the LLM, \( W_{i} \) represents all values in the \( i \)-th row of the weight matrix, and the superscript of \( W \) specifies its associated data type. 

To achieve quantization, we first determine the maximum absolute value of the weights in each channel, denoted as \( \max\{|W_{i}|\} \). Next, based on the selected quantization bit-width \( n \), a specific scaling factor \( s_{i} \) is computed for each channel. This scaling factor serves to map the range of floating-point weights to the integer range required for quantization. Finally, the floating-point weights of each channel are divided by their respective scaling factors and rounded to the nearest integer.  
Through these steps, the weights in each channel are transformed into integer form, completing the quantization process.

\section{Experiments}

\begin{table*}[ht]
\centering
\caption{Zero-shot QA task results of quantized Llama models. $\uparrow$ denotes the larger the better. Bold values denote the best performance scores.}

\scalebox{1.15}{

\begin{tabular}{cccccccccc}
\toprule
\textbf{Model}                        & \textbf{Method}                              & \textbf{Avg. Bits} & \textbf{PIQA}  & \textbf{ARC-e} & \textbf{ARC-c} & \textbf{BoolQ} & \textbf{HellaSwag} & \textbf{WinoGrande} & \textbf{Avg.} $(\uparrow)$   \\ \midrule
\multirow{6}{*}{\textbf{Llama-2-7B}}  & \multirow{3}{*}{\textbf{LWQ}}                & \textbf{7}   & 78.02 & 67.09 & 39.59 & 71.96 & 56.76 & 66.54 & 63.33 \\
                                      &                                              & \textbf{6}   & 77.48 & 65.40 & 39.16 & 67.89 & 56.77 & 66.31 & 62.17          \\
                                      &                                              & \textbf{5}   & 76.66 & 64.65 & 38.57 & 65.57 & 55.41 & 65.74 & 61.10          \\ \cmidrule(l){2-10} 
                                      & \multirow{3}{*}{\textbf{LSAQ (Ours)}}        & \textbf{7}   & 78.02 & 67.09 & 39.59 & 71.96 & 56.76 & 66.54 & {63.33} \\
                                      &                                              & \textbf{6}   & 77.29 & 65.62 & 39.68 & 69.69 & 56.99 & 65.90 & \textbf{62.53} \\
                                      &                                              & \textbf{5}   & 77.15 & 64.74 & 39.19 & 65.99 & 55.55 & 65.59 & \textbf{61.37} \\ \midrule
\multirow{6}{*}{\textbf{Llama-2-13B}} & \multirow{3}{*}{\textbf{LWQ}}                & \textbf{7}   & 78.94 & 72.26 & 45.48 & 71.90 & 59.73 & 68.90 & 66.20          \\
                                      &                                              & \textbf{6}   & 78.78 & 71.94 & 44.45 & 71.90 & 59.08 & 68.51 & 65.78          \\
                                      &                                              & \textbf{5}   & 78.02 & 70.59 & 43.69 & 71.44 & 58.37 & 68.20 & 65.05          \\ \cmidrule(l){2-10} 
                                      & \multirow{3}{*}{\textbf{LSAQ (Ours)}}        & \textbf{7}   & 79.00 & 72.73 & 45.73 & 72.51 & 59.78 & 69.22 & \textbf{66.50} \\
                                      &                                              & \textbf{6}   & 78.85 & 72.32 & 45.52 & 71.35 & 59.29 & 68.98 & \textbf{66.05} \\
                                      &                                              & \textbf{5}   & 78.56 & 70.88 & 43.26 & 71.28 & 58.55 & 68.82 & \textbf{65.23} \\ \midrule
\multirow{6}{*}{\textbf{Llama-3-8B}}  & \multirow{3}{*}{\textbf{LWQ}}                & \textbf{7}   & 79.02 & 79.46 & 49.91 & 81.50 & 59.13 & 73.88 & 70.48          \\
                                      &                                              & \textbf{6}   & 78.67 & 78.37 & 48.81 & 80.06 & 58.83 & 73.22 & 69.66          \\
                                      &                                              & \textbf{5}   & 78.40 & 74.92 & 45.56 & 75.93 & 57.85 & 72.85 & 67.59          \\ \cmidrule(l){2-10} 
                                      & \multirow{3}{*}{\textbf{LSAQ (Ours)}}        & \textbf{7}   & 79.28 & 79.76 & 50.77 & 81.77 & 59.67 & 73.56 & \textbf{70.80} \\
                                      &                                              & \textbf{6}   & 78.85 & 78.76 & 49.02 & 80.86 & 59.14 & 73.48 & \textbf{70.02} \\
                                      &                                              & \textbf{5}   & 78.63 & 77.02 & 47.27 & 79.94 & 58.24 & 73.48 & \textbf{69.10} \\ \bottomrule
\end{tabular}

}

\label{table:zeroshot}
\end{table*}

\subsection{Experimental Setup}
\subsubsection{LLMs}
To verify the feasibility of our method, we conducted experiments on state-of-the-art open-source LLMs, including Llama-2-7B, Llama-2-13B, and Llama-3-8B, developed by Meta AI. These LLMs are based on the decoder-only Transformer architecture. Due to their extensive pre-training on large-scale language data, they demonstrate exceptional performance across a variety of natural language processing tasks.

\subsubsection{Benchmarks}
To comprehensively evaluate the performance of LLMs after quantization, we adopted a dual evaluation strategy. 
Firstly, we selected six zero-shot tasks to evaluate whether the core capabilities of the LLMs are preserved after quantization:

\begin{compactitem}
    \item PIQA \cite{bisk2020piqa}: A physical commonsense reasoning dataset that tests a model's ability to choose appropriate solutions in everyday physical scenarios by providing a physical goal and two possible solutions. It emphasizes understanding the properties and manipulation of objects and contains over 16,000 question-answer pairs to assess the model's physical commonsense reasoning ability.
    \item ARC-e, ARC-c: These two datasets represent the easy and challenging subsets of the ARC \cite{clark2018think} (AI2 Reasoning Challenge). ARC is a scientific reasoning dataset comprising approximately 8,000 questions and includes a 14 million scientific facts corpus to support answering these questions.
    \item BoolQ \cite{clark2019boolq}: A question-answering task where the model must predict whether a passage contains the answer to a specific question. These questions require a "yes" or "no" response based on the given passage.
    \item HellaSwag \cite{zellers2019hellaswag}: A dataset designed to test the commonsense reasoning ability of models. It provides a series of scenarios requiring the model to select the most appropriate outcome from multiple options. This dataset demands complex reasoning based on context rather than simple word or phrase matching.
    \item WinoGrande \cite{sakaguchi2021winogrande}: A large-scale commonsense reasoning dataset, based on the design of the Winograd Schema Challenge, which uses crowdsourcing and bias reduction algorithms to generate 44,000 problems. It aims to more accurately assess the common sense reasoning capabilities of artificial intelligence by increasing the scale and difficulty of the problems.

\end{compactitem}

By testing the accuracy of LLMs on the aforementioned zero-shot tasks, we can assess the reasoning and generalization capabilities of the quantized LLMs during their usage.

Secondly, we have assessed the perplexity (PPL) of quantized LLMs on the WikiText2 \cite{merity2016pointer} dataset. This dataset is derived from Wikipedia articles and encompasses a wealth of encyclopedic knowledge. By measuring the PPL of LLMs in language generation tasks across various levels of quantization precision, we are able to accurately evaluate the LLMs' predictive capabilities on the test data.

\subsubsection{Baseline}
In the experiment, we selected Layer-Wise Quantization (hereinafter referred to as LWQ) as the baseline method. The LWQ technique aligns with our quantization strategy in terms of granularity, since they both apply differential quantization based on the importance of each layer. The LWQ method evaluates layer importance using two different metrics: one is the cosine similarity between the layer's input and output, and the other is the number of weights in the layer that are significantly greater than the average value. 

Experimental results from the LWQ paper indicate that using cosine similarity as the basis for quantization leads to better performance for the Llama series models. On the basis of this consideration, we decided to use cosine similarity as the layer importance metric for LWQ in formulating a quantization strategy. Since this method is independent of the underlying quantization technique, we will test it using the same quantization approach as ours.

\subsection{Main Results}\label{sec:main_results}

Tables \ref{table:zeroshot} and \ref{table:ppl} present the detailed results of our experiments. To comprehensively evaluate the application of LSAQ quantization technology to LLMs, we tested the zero-shot task accuracy and perplexity of quantized Llama-2-7B/13B and Llama-3-8B models. In these experiments, INT4 quantization was applied to 25\%, 50\%, and 75\% of LLMs' layers, while the remaining layers used INT8 quantization. This quantization scheme reduced the average bit-width of LLMs to 7 bits, 6 bits, and 5 bits, corresponding to different quantization ratios.

\subsubsection{Results on zero-shot tasks}
Table \ref{table:zeroshot} provides a detailed breakdown of the accuracy and overall average scores of various quantized LLMs across multiple zero-shot question-answering tasks. 

\begin{table}[ht]
    \renewcommand{\arraystretch}{1.5}
    \setlength{\tabcolsep}{6pt}
\centering
\caption{The first 25\% layer numbers of Llama-2-7B under different layer importance indicators.}
\begin{tabular}{ccc}
\toprule
\textbf{Model}                       & \textbf{Importance Indicator} & \textbf{The First 25\% Layers}                  \\ \midrule
\multirow{2}{*}{\textbf{Llama-2-7B}} & \multirow{1}{*}{\textbf{Cosine}}    & \multirow{1}{*}{27, 26, 28, 25, 24, 29, 23, 22} \\
                                     % &                                     &                                                 \\
                                     & \multirow{1}{*}{\textbf{Jaccard (Ours)}}   & \multirow{1}{*}{27, 26, 28, 24, 23, 25, 22, 29} \\ \bottomrule
                                     % &                                     &                                                 \\ 
\end{tabular}
\label{table:layer_num}
\end{table}

However, there is a special case in the experimental results. On the Llama-2-7B model, the performance of the two quantization methods is exactly the same under the condition of 7-bit quantization. \textcolor{black}{We attribute it to the fact that although there are differences in determining the order of the first 25\% least important layers with two indicators,} they ultimately reach a consensus at the overall level, as shown in Table \ref{table:layer_num}. Consequently, both methods exhibit a high degree of consistency across all performance metrics. This consistency is reflected in both the zero-shot evaluation experiments and the subsequent perplexity experiments.

From the experimental data, it can be observed that the LSAQ quantized LLM generally achieves superior performance in these zero-shot tasks. Except for the 25\% quantization case of Llama-2-7B (where both methods show high consistency), the LSAQ method outperforms the LWQ method in terms of average accuracy across all tasks and has an advantage in accuracy for \textbf{87.5\%} of individual tasks. Notably, for the ARC-e and HellaSwag tasks, our method achieves higher accuracy under a wide range of conditions. Moreover, the LSAQ method is able to maintain a high level of accuracy even at lower bit numbers.

\subsubsection{Results on perplexity}

The PPL is a critical metric in the field of NLP used to evaluate the performance of LLMs. It measures the LLM's ability to predict a set of sample data, particularly in assessing the LLM's predictive accuracy and generalization capability. A lower perplexity value indicates higher prediction accuracy, thereby reflecting the impact of quantization on LLM performance.

Table \ref{table:ppl} demonstrates the impact of different quantization methods on the perplexity of three LLMs at various quantization bit widths. The experimental results indicate that, except for the condition of $<$Llama-3-8B, 7 bits$>$ where the LWQ quantization slightly outperforms LSAQ, the LLM quantized by LSAQ exhibits lower perplexity on the WikiText2 dataset in all the remaining cases.

\begin{table}[ht]
    \renewcommand{\arraystretch}{1.5}
    \setlength{\tabcolsep}{6pt}
\centering
\caption{Perplexity results of quantized Llama models on the WikiText2 dataset.}
\begin{tabular}{ccccccc}
\toprule
\textbf{Model}                        & \textbf{Method} & \textbf{8-bits}        & \textbf{7-bits} & \textbf{6-bits} & \textbf{5-bits} & \textbf{4-bits}         \\ \midrule
\multirow{2}{*}{\textbf{Llama-2-7B}}  & \textbf{LWQ}    & \multirow{2}{*}{5.476} & 5.771           & 6.156           & 6.396           & \multirow{2}{*}{6.919}  \\
                                      & \textbf{LSAQ}   &                        & \textbf{5.771}  & \textbf{6.064}  & \textbf{6.325}  &                         \\ \midrule
\multirow{2}{*}{\textbf{Llama-2-13B}} & \textbf{LWQ}    & \multirow{2}{*}{4.886} & 4.958           & 5.074           & 5.200           & \multirow{2}{*}{5.403}  \\
                                      & \textbf{LSAQ}   &                        & \textbf{4.940}  & \textbf{5.057}  & \textbf{5.162}  &                         \\ \midrule
\multirow{2}{*}{\textbf{Llama-3-8B}}  & \textbf{LWQ}    & \multirow{2}{*}{6.143} & \textbf{6.443}  & 6.904           & 7.386           & \multirow{2}{*}{10.530} \\
                                      & \textbf{LSAQ}   &                        & 6.508           & \textbf{6.887}  & \textbf{7.331}  &                         \\ \bottomrule
\end{tabular}
\label{table:ppl}
\end{table}

\subsubsection{Analysis of results}
By analyzing the experimental data of the quantized LLM on zero-shot tasks and perplexity, it can be demonstrated that evaluating the layer importance of LLMs by constructing the top-\(k\) token sets corresponding to the input and output of each layer and calculating the Jaccard similarity of these sets is more effective than using cosine similarity. This evaluation approach adeptly captures the interplay between each layer and semantic information. Consequently, it facilitates a more nuanced and accurate appraisal of the significance of each layer.

This finding reveals that compared to LWQ, the LSAQ quantization method is more effective in maintaining LLM accuracy, as it loses less critical information during the quantization process. Consequently, this suggests that LSAQ has potential advantages in the field of LLM compression and deployment on edge devices.

\subsection{Quantized LLM Deployment}

In addition to quantization, dynamic deployment of LLMs on edge devices is also one of the core capabilities of LSAQ. To investigate whether LSAQ meets the criteria for the successful deployment of LLMs on edge devices, we performed tests on the memory requirements for loading LLM weights of the quantized Llama-2-7B/13B and Llama-3-8B. Besides, we examined the quantization strategies provided by our system for the Llama-2-7B model under various memory constraints to assess whether they are compliant with the requirements.

\begin{figure}[ht]
    \centering
    \includegraphics[width=1\linewidth]{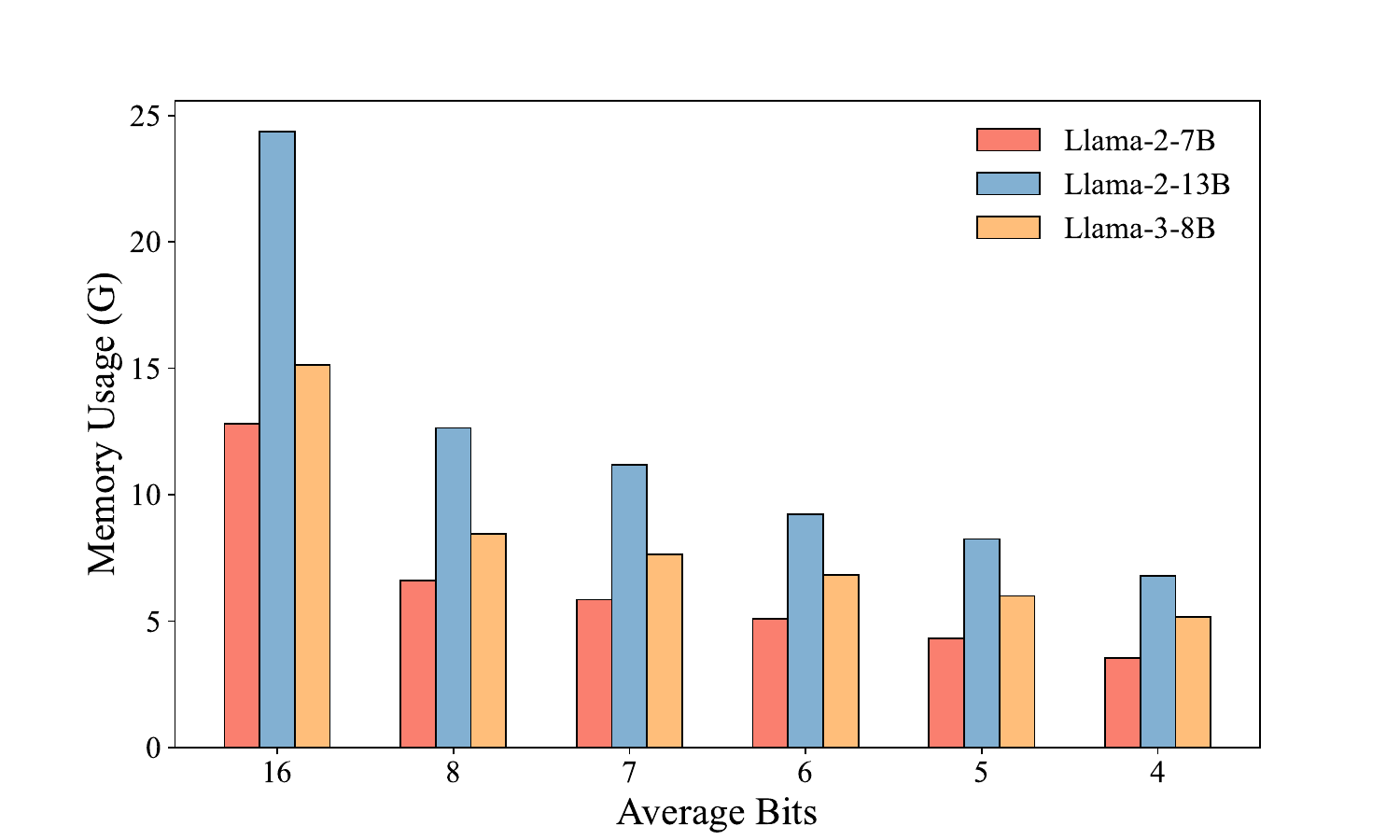}
    \caption{Memory usage of quantized model}
    \label{fig:gpu_use}
\end{figure}

Figure \ref{fig:gpu_use} demonstrates the amount of memory required to load the weights of three LLMs at different quantization precisions. When these LLMs operate at FP16 precision, they require approximately 12.82GB, 24.36GB, and 15.14GB of memory, respectively. With the reduction of the average number of bits after quantization, the memory occupancy of all LLMs shows significant decreases. At their lowest, the memory requirements can be reduced to about 3.56GB, 6.79GB, and 5.18GB, respectively, which allows these LLMs to be easily deployed on mainstream graphics cards currently available on the market.

\begin{table}[ht]
    \renewcommand{\arraystretch}{1.2}
    \setlength{\tabcolsep}{6pt}
\centering
\caption{Quantization strategies for Llama-2-7B under various memory conditions.}
\begin{tabular}{c|cccc}
\toprule
\textbf{Memory} & \textbf{FP16 Layer} & \textbf{INT8 Layer} & \textbf{INT4 Layer} & \textbf{Avg. Bits} \\ \midrule
\textbf{16G}    & 32                  & 0                   & 0                   & 16            \\
\textbf{12G}    & 0                   & 32                  & 0                   & 8             \\
\textbf{8G}     & 0                   & 32                  & 0                   & 8             \\
\textbf{6G}     & 0                   & 32                  & 10                  & 6.75          \\
\textbf{4G}     & 0                   & 1                   & 31                  & 4.125         \\ \bottomrule
\end{tabular}
\label{table:strategies}
\end{table}

During the LLM inference process, intermediate computational results are generated, which also occupy a certain amount of memory. Therefore, when designing the quantization strategy, we reserved a portion of the storage for its inference based on the requirements of LLMs of different sizes.

Table \ref{table:strategies} illustrates the actual quantization strategies employed by LSAQ for deploying Llama-2-7B under various memory configurations. For instance, if a GPU has 16GB of available memory, it is entirely feasible to deploy Llama-2-7B directly on the GPU at FP16 precision without the need for quantization. However, when the GPU has only 6GB of available memory, it becomes necessary to quantize the least important 10 out of the 32 layers of Llama-2-7B to INT4 precision, while the remaining 22 layers are quantized to INT8 precision, thereby facilitating the deployment of the LLM.

\section{Conclusion}

This paper introduces LSAQ, a Layer-Specific Adaptive Quantization and dynamic deployment system for LLMs. LSAQ adjusts quantization strategies on hardware platforms with varying resource sizes by establishing a novel layer importance assessment mechanism, thereby significantly reducing the LLMs' storage requirements while maintaining LLM performance as much as possible. Experimental results demonstrate that the accuracy of the LSAQ-quantized LLM on zero-shot tasks and the perplexity of the LLM are superior to existing quantization methods of the same granularity, even at lower bit widths, thus maintaining high performance. This indicates that LSAQ has potential advantages in LLM compression and on-device LLM deployment. By employing intelligent adaptive quantization methods, LSAQ can efficiently deploy quantized LLMs in resource-limited environments, providing a new perspective and approach for the deployment of LLMs on edge devices.

\section{Acknowledgement}
This work was supported by the National Key Research and Development Program under Grant 2024YFB4506200, the Science and Technology Innovation Program of Hunan Province under Grant 2024RC1048, and the National Key Laboratory Foundation Project under Grant 2024-KJWPDL-14.

% \nocite{*}
\bibliographystyle{IEEEtran}
\bibliography{refs}

\begin{thebibliography}{10}
\providecommand{\url}[1]{#1}
\csname url@samestyle\endcsname
\providecommand{\newblock}{\relax}
\providecommand{\bibinfo}[2]{#2}
\providecommand{\BIBentrySTDinterwordspacing}{\spaceskip=0pt\relax}
\providecommand{\BIBentryALTinterwordstretchfactor}{4}
\providecommand{\BIBentryALTinterwordspacing}{\spaceskip=\fontdimen2\font plus
\BIBentryALTinterwordstretchfactor\fontdimen3\font minus \fontdimen4\font\relax}
\providecommand{\BIBforeignlanguage}[2]{{%
\expandafter\ifx\csname l@#1\endcsname\relax
\typeout{** WARNING: IEEEtran.bst: No hyphenation pattern has been}%
\typeout{** loaded for the language `#1'. Using the pattern for}%
\typeout{** the default language instead.}%
\else
\language=\csname l@#1\endcsname
\fi
#2}}
\providecommand{\BIBdecl}{\relax}
\BIBdecl

\bibitem{qin2024largelanguagemodelsmeet}
\BIBentryALTinterwordspacing
L.~Qin, Q.~Chen, X.~Feng, Y.~Wu, Y.~Zhang, Y.~Li, M.~Li, W.~Che, and P.~S. Yu, ``Large language models meet nlp: A survey,'' 2024. [Online]. Available: \url{https://arxiv.org/abs/2405.12819}
\BIBentrySTDinterwordspacing

\bibitem{jiang2024surveylargelanguagemodels}
\BIBentryALTinterwordspacing
J.~Jiang, F.~Wang, J.~Shen, S.~Kim, and S.~Kim, ``A survey on large language models for code generation,'' 2024. [Online]. Available: \url{https://arxiv.org/abs/2406.00515}
\BIBentrySTDinterwordspacing

\bibitem{li2024modeleditingllms4codefar}
\BIBentryALTinterwordspacing
X.~Li, S.~Wang, S.~Li, J.~Ma, J.~Yu, X.~Liu, J.~Wang, B.~Ji, and W.~Zhang, ``Model editing for llms4code: How far are we?'' 2024. [Online]. Available: \url{https://arxiv.org/abs/2411.06638}
\BIBentrySTDinterwordspacing

\bibitem{zhao2024revolutionizing}
H.~Zhao, Z.~Liu, Z.~Wu, Y.~Li, T.~Yang, P.~Shu, S.~Xu, H.~Dai, L.~Zhao, G.~Mai \emph{et~al.}, ``Revolutionizing finance with llms: An overview of applications and insights,'' \emph{arXiv preprint arXiv:2401.11641}, 2024.

\bibitem{wang2024large}
S.~Wang, T.~Xu, H.~Li, C.~Zhang, J.~Liang, J.~Tang, P.~S. Yu, and Q.~Wen, ``Large language models for education: A survey and outlook,'' \emph{arXiv preprint arXiv:2403.18105}, 2024.

\bibitem{lyu2023llm}
H.~Lyu, S.~Jiang, H.~Zeng, Y.~Xia, Q.~Wang, S.~Zhang, R.~Chen, C.~Leung, J.~Tang, and J.~Luo, ``Llm-rec: Personalized recommendation via prompting large language models,'' \emph{arXiv preprint arXiv:2307.15780}, 2023.

\bibitem{touvron2023llama}
H.~Touvron, L.~Martin, K.~Stone, P.~Albert, A.~Almahairi, Y.~Babaei, N.~Bashlykov, S.~Batra, P.~Bhargava, S.~Bhosale \emph{et~al.}, ``Llama 2: Open foundation and fine-tuned chat models,'' \emph{arXiv preprint arXiv:2307.09288}, 2023.

\bibitem{frantar2022gptq}
E.~Frantar, S.~Ashkboos, T.~Hoefler, and D.~Alistarh, ``Gptq: Accurate post-training quantization for generative pre-trained transformers,'' \emph{arXiv preprint arXiv:2210.17323}, 2022.

\bibitem{dettmers2022llmint88bitmatrixmultiplication}
\BIBentryALTinterwordspacing
T.~Dettmers, M.~Lewis, Y.~Belkada, and L.~Zettlemoyer, ``Llm.int8(): 8-bit matrix multiplication for transformers at scale,'' 2022. [Online]. Available: \url{https://arxiv.org/abs/2208.07339}
\BIBentrySTDinterwordspacing

\bibitem{lin2024duquantdistributingoutliersdual}
\BIBentryALTinterwordspacing
H.~Lin, H.~Xu, Y.~Wu, J.~Cui, Y.~Zhang, L.~Mou, L.~Song, Z.~Sun, and Y.~Wei, ``Duquant: Distributing outliers via dual transformation makes stronger quantized llms,'' 2024. [Online]. Available: \url{https://arxiv.org/abs/2406.01721}
\BIBentrySTDinterwordspacing

\bibitem{ma2023llm}
X.~Ma, G.~Fang, and X.~Wang, ``Llm-pruner: On the structural pruning of large language models,'' \emph{Advances in neural information processing systems}, vol.~36, pp. 21\,702--21\,720, 2023.

\bibitem{frantar2023sparsegpt}
E.~Frantar and D.~Alistarh, ``Sparsegpt: Massive language models can be accurately pruned in one-shot,'' in \emph{International Conference on Machine Learning}.\hskip 1em plus 0.5em minus 0.4em\relax PMLR, 2023, pp. 10\,323--10\,337.

\bibitem{yang2024lacolargelanguagemodel}
\BIBentryALTinterwordspacing
Y.~Yang, Z.~Cao, and H.~Zhao, ``Laco: Large language model pruning via layer collapse,'' 2024. [Online]. Available: \url{https://arxiv.org/abs/2402.11187}
\BIBentrySTDinterwordspacing

\bibitem{gu2024minillm}
Y.~Gu, L.~Dong, F.~Wei, and M.~Huang, ``Minillm: Knowledge distillation of large language models,'' in \emph{The Twelfth International Conference on Learning Representations}, 2024.

\bibitem{huang2022context}
Y.~Huang, Y.~Chen, Z.~Yu, and K.~McKeown, ``In-context learning distillation: Transferring few-shot learning ability of pre-trained language models,'' \emph{arXiv preprint arXiv:2212.10670}, 2022.

\bibitem{xu2023tensorgpt}
M.~Xu, Y.~L. Xu, and D.~P. Mandic, ``Tensorgpt: Efficient compression of the embedding layer in llms based on the tensor-train decomposition,'' \emph{arXiv preprint arXiv:2307.00526}, 2023.

\bibitem{shao2024omniquantomnidirectionallycalibratedquantization}
\BIBentryALTinterwordspacing
W.~Shao, M.~Chen, Z.~Zhang, P.~Xu, L.~Zhao, Z.~Li, K.~Zhang, P.~Gao, Y.~Qiao, and P.~Luo, ``Omniquant: Omnidirectionally calibrated quantization for large language models,'' 2024. [Online]. Available: \url{https://arxiv.org/abs/2308.13137}
\BIBentrySTDinterwordspacing

\bibitem{men2024shortgpt}
X.~Men, M.~Xu, Q.~Zhang, B.~Wang, H.~Lin, Y.~Lu, X.~Han, and W.~Chen, ``Shortgpt: Layers in large language models are more redundant than you expect,'' \emph{arXiv preprint arXiv:2403.03853}, 2024.

\bibitem{dumitru2024layer}
R.-G. Dumitru, V.~Yadav, R.~Maheshwary, P.-I. Clotan, S.~T. Madhusudhan, and M.~Surdeanu, ``Layer-wise quantization: A pragmatic and effective method for quantizing llms beyond integer bit-levels,'' \emph{arXiv preprint arXiv:2406.17415}, 2024.

\bibitem{merity2016pointer}
S.~Merity, C.~Xiong, J.~Bradbury, and R.~Socher, ``Pointer sentinel mixture models,'' \emph{arXiv preprint arXiv:1609.07843}, 2016.

\bibitem{yao2022zeroquant}
Z.~Yao, R.~Yazdani~Aminabadi, M.~Zhang, X.~Wu, C.~Li, and Y.~He, ``Zeroquant: Efficient and affordable post-training quantization for large-scale transformers,'' \emph{Advances in Neural Information Processing Systems}, vol.~35, pp. 27\,168--27\,183, 2022.

\bibitem{xiao2023smoothquant}
G.~Xiao, J.~Lin, M.~Seznec, H.~Wu, J.~Demouth, and S.~Han, ``Smoothquant: Accurate and efficient post-training quantization for large language models,'' in \emph{International Conference on Machine Learning}.\hskip 1em plus 0.5em minus 0.4em\relax PMLR, 2023, pp. 38\,087--38\,099.

\bibitem{lin2024awq}
J.~Lin, J.~Tang, H.~Tang, S.~Yang, W.-M. Chen, W.-C. Wang, G.~Xiao, X.~Dang, C.~Gan, and S.~Han, ``Awq: Activation-aware weight quantization for on-device llm compression and acceleration,'' \emph{Proceedings of Machine Learning and Systems}, vol.~6, pp. 87--100, 2024.

\bibitem{dumitru2024change}
R.-G. Dumitru, P.-I. Clotan, V.~Yadav, D.~Peteleaza, and M.~Surdeanu, ``Change is the only constant: Dynamic llm slicing based on layer redundancy,'' \emph{arXiv preprint arXiv:2411.03513}, 2024.

\bibitem{li2024pmet}
X.~Li, S.~Li, S.~Song, J.~Yang, J.~Ma, and J.~Yu, ``Pmet: Precise model editing in a transformer,'' in \emph{Proceedings of the AAAI Conference on Artificial Intelligence}, vol.~38, no.~17, 2024, pp. 18\,564--18\,572.

\bibitem{niwattanakul2013using}
S.~Niwattanakul, J.~Singthongchai, E.~Naenudorn, and S.~Wanapu, ``Using of jaccard coefficient for keywords similarity,'' in \emph{Proceedings of the international multiconference of engineers and computer scientists}, vol.~1, no.~6, 2013, pp. 380--384.

\bibitem{dubey2024llama}
A.~Dubey, A.~Jauhri, A.~Pandey, A.~Kadian, A.~Al-Dahle, A.~Letman, A.~Mathur, A.~Schelten, A.~Yang, A.~Fan \emph{et~al.}, ``The llama 3 herd of models,'' \emph{arXiv preprint arXiv:2407.21783}, 2024.

\bibitem{bisk2020piqa}
Y.~Bisk, R.~Zellers, J.~Gao, Y.~Choi \emph{et~al.}, ``Piqa: Reasoning about physical commonsense in natural language,'' in \emph{Proceedings of the AAAI conference on artificial intelligence}, vol.~34, no.~05, 2020, pp. 7432--7439.

\bibitem{clark2018think}
P.~Clark, I.~Cowhey, O.~Etzioni, T.~Khot, A.~Sabharwal, C.~Schoenick, and O.~Tafjord, ``Think you have solved question answering? try arc, the ai2 reasoning challenge,'' \emph{arXiv preprint arXiv:1803.05457}, 2018.

\bibitem{clark2019boolq}
C.~Clark, K.~Lee, M.-W. Chang, T.~Kwiatkowski, M.~Collins, and K.~Toutanova, ``Boolq: Exploring the surprising difficulty of natural yes/no questions,'' \emph{arXiv preprint arXiv:1905.10044}, 2019.

\bibitem{zellers2019hellaswag}
R.~Zellers, A.~Holtzman, Y.~Bisk, A.~Farhadi, and Y.~Choi, ``Hellaswag: Can a machine really finish your sentence?'' \emph{arXiv preprint arXiv:1905.07830}, 2019.

\bibitem{sakaguchi2021winogrande}
K.~Sakaguchi, R.~L. Bras, C.~Bhagavatula, and Y.~Choi, ``Winogrande: An adversarial winograd schema challenge at scale,'' \emph{Communications of the ACM}, vol.~64, no.~9, pp. 99--106, 2021.

\end{thebibliography}
\end{document}